\documentclass[runningheads]{llncs}

\usepackage[mobile]{eccv}
\usepackage{abbrv}
\usepackage{graphicx}
\usepackage{booktabs}
\usepackage{comment}

\usepackage[accsupp]{axessibility}  

\def\smalName{SMAL$^{+}$\xspace}
\def\methodName{AWOL\xspace}
\def\rnvpName{Real-NVP\xspace}

\usepackage[breaklinks,colorlinks]{hyperref}
\usepackage{orcidlink}

\begin{document}
\title{AWOL: Analysis WithOut synthesis \\ using Language}

\titlerunning{AWOL}

\author{Silvia Zuffi\inst{1} \and
Michael J. Black\inst{2}}

\authorrunning{S.~Zuffi et al.}

\institute{IMATI-CNR, Milan, Italy \\
\email{silvia@mi.imati.cnr.it}\and
Max~Planck~Institute~for~Intelligent~Systems, T\"{u}bingen, Germany \\
\email{black@tue.mpg.de}}

\maketitle
\begin{figure}[h]
\vspace{-1cm}
    \centering
    \includegraphics[width=\textwidth]{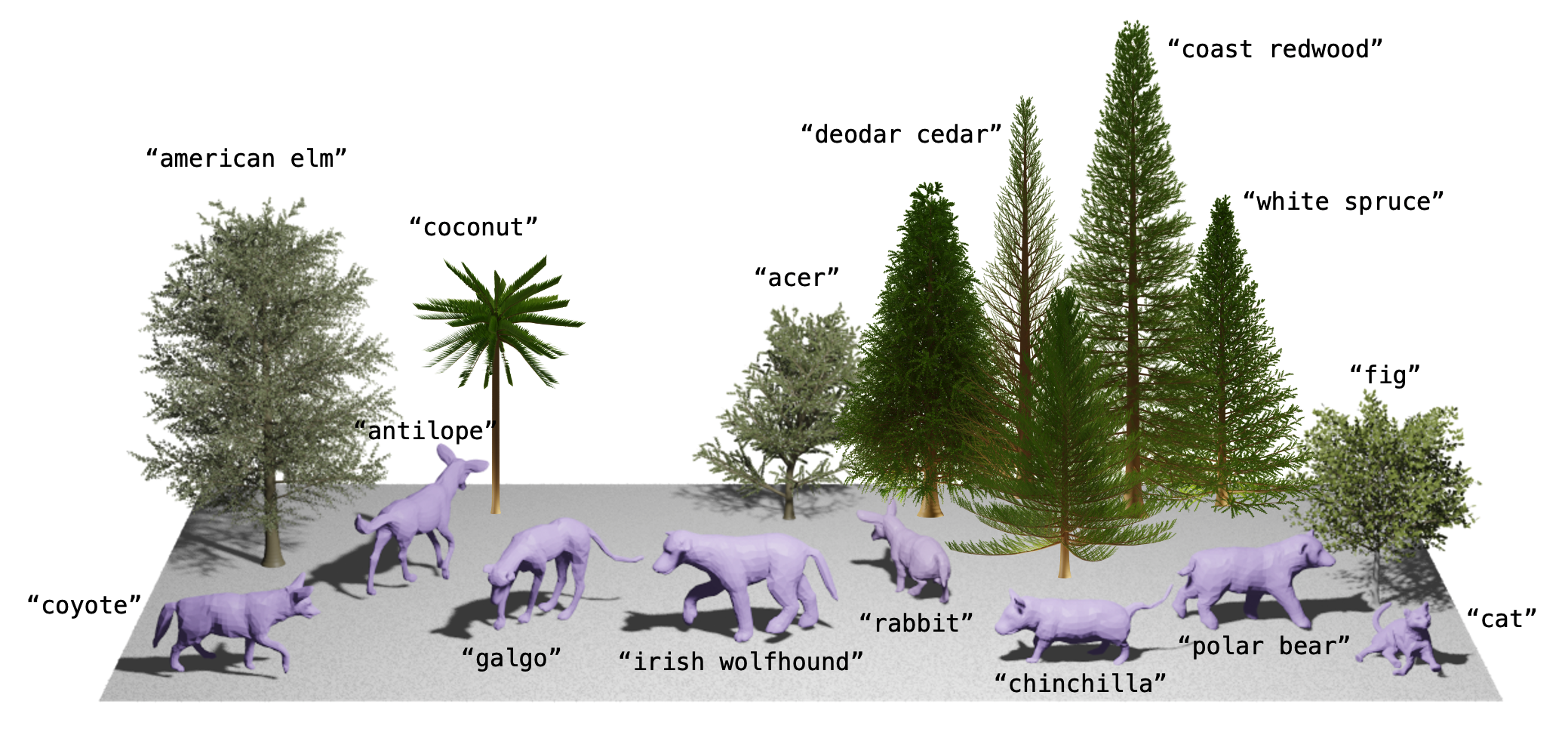}
    \caption{Generated trees and animals. \methodName learns to generate animals and trees from text and images. We show examples for text-generated tree and animal species not seen during training (except for the cat).}
    \label{fig:teaser_2}
\end{figure}
\vspace{-4mm}

\begin{abstract}
\vspace{-1cm}
Many classical parametric 3D shape models exist, but creating novel shapes with such models requires expert knowledge of their parameters.
For example, imagine creating a specific type of tree using procedural graphics or a new kind of animal from a statistical shape model.
Our key idea is to leverage language to control such existing models to produce novel shapes.
This involves learning a mapping between the latent space of a vision-language model and the parameter space of the 3D model, which we do using a small set of shape and text pairs. 
Our hypothesis is that mapping from language to parameters allows us to generate parameters for objects that were never seen during training.
If the mapping between language and parameters is sufficiently smooth, then interpolation or generalization in language should translate appropriately into novel 3D shapes.
We test our approach with two very different types of parametric shape models (quadrupeds and arboreal trees).
We use a learned statistical shape model of quadrupeds and show that we can use text to generate new animals not present during training.
In particular, we demonstrate state-of-the-art shape estimation of 3D dogs.
This work also constitutes the first language-driven method for generating 3D trees.
Finally, embedding images in the CLIP latent space enables us to generate animals and trees directly from images.
\end{abstract}

\section{Introduction}
\label{sec:intro}

We address the problem of generating new, realistic samples from various 3D shape models using language.
The key idea is to relate language (e.g.~names of dog breeds or types of trees) to the model's parameters and then leverage language to generate shapes that were never seen during training.
To make this possible, we leverage  the shared latent space of large vision-language foundation models (VLM), like CLIP (Contrastive Language-Image Pretraining) \cite{pmlr-v139-radford21a_clip}.
Such models relate how objects appear in images to how we describe them with language. 
Since how objects appear is related to their 3D shape, we can assume that CLIP implicitly also relates object shape with language.
Since models like CLIP are learned from large data corpora, VLM latent spaces are rich and dense; in other words, they know a lot about objects and their shape, but not explicitly.
Given a small training set, we learn a mapping between the CLIP space and the shape parameters of various models.
Finally, our central hypothesis is that the CLIP space is well-behaved such that interpolation or extrapolation in this space produce appropriate interpolation or extrapolation of the associated shape parameters.
This allows us to exploit the general knowledge of a VLM to control the parameters of the shape model to produce novel shapes outside its training set.
We test our hypothesis using two very diverse object classes, animals and trees, that use two very different generation processes. 
For animals, we use an analytic, statistical, parametric shape model, named \smalName, that we introduce here as a new, extended version of previous models \cite{Zuffi:CVPR:2017, bite2023rueegg, li2021hsmal}.
For trees, we use a procedural, non-differentiable, tree generator implemented as a Blender add-on \cite{treegen}; this is very different from \smalName.  
Trees are an interesting case because they are composed by thin structures (branches) and thin surfaces (leaves) that cannot easily be fit with the 3D implicit representations used in many current text-to-3D solutions.
With our method, named \methodName, we generate trees and animals that are unseen during training and that are expressed as triangular meshes, thus supporting easy rendering and animation in graphics engines; see Fig.~\ref{fig:teaser_2}.

There is growing interest in generating 3D content with easy-to-use tools. An abundance of methods have been proposed to create 3D assets from simple text prompts, or single images \cite{xu2023_dream3d, poole2022_dreamfusion, cheng2023_sdfusion, JainMBAP22}.  Such methods are able to generate compelling rigid objects, with realistic appearance. 
Such models do not, however, produce articulated objects that are rigged for animation.
With \methodName, we obtain animal models that share the same skeleton and mesh topology. This is important: a standardized 3D generation would allow easy motion transfer and facilitate analysis, promoting the application of 3D computer vision methods (i.e. 3D model-based articulated motion estimation)  
to the animal research and conservation fields.

Existing 3D parametric shape models for articulated subjects, like SMPL \cite{SMPL:2015}, for humans, or SMAL \cite{zuffi2019three}, for animals, are generative models for body shape, and consequently they are widely used to create 3D avatars, either by sampling the generative model, or by aligning the model to data \cite{Bogo:ECCV:2016, pavlakos2022multishot, pavlakos2022sitcoms3D, goel2023humans, mueller2023buddi, DSR:ICCV:2021, Kocabas_SPEC_2021, ROMP:ICCV:2021, Kocabas_PARE_2021, PIXIE:3DV:2021, Shapy:2022, tripathi2023ipman, Sun:CVPR:2023, Zuffi:CVPR:2018,zuffi2019three}. Alignment is made possible by the differentiable nature of the models, which support reconstruction through the analysis-by-synthesis paradigm.
While the SMPL model can arguably represent a large portion of the world population, given its large training set and uniqueness of the human species, the SMAL model has been trained on a small set of quadrupeds to represent animals from $5$ different families (canine, equine, bovine, hippopotamids, and feline). As such, naively sampling the model shape space can produce non-existing animals that are often a mixture of more species. Sampling with family-specific shape priors (i.e. Gaussian distributions centered at the family mean shape variables) allows generating instances with realistic shape. However, as also illustrated in the paper \cite{Zuffi:CVPR:2017}, when aligned to data, the SMAL model can broadly represent species that are not present in its training set, for example representing a boar with a mane borrowed by lions, a long mouth from hippos, and bulky body from cows. The question then is: how can we generate animal species that are not in one of the five SMAL families without analysis-by-synthesis?
The question is of broader application, as it regards the possibility of generalizing the generation of 3D assets given parametric models defined on a small set of samples.  
Identifying the manifold of realistic samples can be difficult: some regions of the space can correspond to shapes not seen during training, but realistic, while other regions can correspond to non-existing class instances.   
There is therefore a problem of realistic interpolation for data generation.
In addition, shape models based on continuous latent spaces do not offer extrapolation capabilities, as in general their dimensions do not correspond to semantic deformations. While space transformations can be applied to identify axes with semantic meaning, this does not address the generalization principle, as how to move along these axes to generate new, realistic samples would still be not defined.
In both the animals and trees models, the set of training samples is scarce. This limits the application of highly flexible generative models that are popular today, i.e. diffusion models. We employ \rnvpName~\cite{Dinh2017realNVP}, a generative model for highly structured data, characterized by a set of explicit transformations defined through a cascade of layers that selectively couple the different dimensions of the input data by means of binary masks. While the model has been used for text-to-3D generation before \cite{sanghi2021clipForge}, using fixed masks, here we show that learning the binary masks improves performance, and, in particular for the animals model, add realistic relative scaling to the predicted shapes. 

In summary, our contributions are: a new 3D parametric shape model for animals, which includes more species than previous models; a method to generate 3D rigged animals from text or images;  a method to generate 3D trees from text or images, which can output a triangular mesh with fine branches and leaves details.

\section{Related Work}
\label{sec:previous}
\paragraph{Text-to-3D}
Our work is related to text-driven model-based 3D content creation systems. 
An early example is BodyTalk \cite{Streuber:SIGGRAPH:2016}, which correlates textual shape attributes with transformed dimensions of the SMPL shape space. 
Semantify \cite{Gralnik_2023_ICCV} also addresses the problem of controlling the SMPL body model with shape attributes, but exploiting CLIP\cite{pmlr-v139-radford21a_clip}. Recent work uses text to control 3D face generation \cite{wu2023describe3d}.
In the past few years, an abundance of methods has addressed the text-driven generation of images \cite{pmlr-v139-ramesh21a, rombach2021highresolution, ramesh2022hierarchical, saharia2022photorealistic, ruiz2022dreambooth, Ge_2023_ICCV}, and more recently 3D objects \cite{hu_clipxplore, JainMBAP22, cheng2023_sdfusion, xu2023_dream3d, sjc2023, Lorraine_2023_ICCV}.        
Training is often based on the similarity between textual queries and rendered 3D shapes when encoded in a joint latent space (i.e. CLIP), with the gradient back propagated through a differentiable renderer. Many methods are thus based on differentiable 3D neural representations, often Neural Radiance Field (NeRF)~\cite{Nerf2021}, with a few mesh-based exceptions \cite{tsalicoglou2024textmesh, lin2023magic3d}. Directly regressing a 3D triplane representation speeds up the text-to-3D generation \cite{li2023instant3d}.  
Recent methods overcome the scarcity of 3D data by exploiting 2D losses. DreamFields generates open-set 3D objects by optimization. The output is a NeRF that is trained by optimizing for rendered views to have high semantic similarity, given the text prompt. The method uses CLIP in synergy with geometric priors. DreamFusion \cite{poole2022_dreamfusion} leverages powerful text-to-image diffusion models (here Imagen \cite{saharia2022photorealistic}) and introduces Score Distillation Sampling (SDS) to exploit diffusion priors as losses for 3D object optimization, an approach also adopted in \cite{sjc2023}.
Our work is related to CLIP-Forge \cite{sanghi2021clipForge}, which trains a normalizing flow network to learn the mapping between the CLIP and the latent space of a 3D shape model, learned over a collection of 3D rigid objects. 

\paragraph{3D Animal Models}
Three-dimensional differentiable articulated shape models have been defined for a few common species. SMAL \cite{Zuffi:CVPR:2017} is a multi-species model that can represent a wide range of quadrupeds. SMALR \cite{Zuffi:CVPR:2018} extends SMAL to capture 3D shapes of animals from a set of images. SMALST \cite{zuffi2019three} learns a 3D model for the Gravy's zebra from images. AVES \cite{wang21aves} learns 3D shape of birds from images, starting from a reference template. hSMAL \cite{li2021hsmal} and D-SMAL \cite{bite2023rueegg} are 3D parametric shape models for horses and dogs, respectively.
Many recent methods do not assume an existing reference template. Lassie \cite{yao2022lassie} and Hi-Lassie \cite{yao2023hi-lassie} create 3D models from a small collection of images. Like SMALR \cite{Zuffi:CVPR:2018}, they require different images with a clear, non occluded view of the animal. Artic3D \cite{yao2023artic3d} supports noisy images. 
Lepard \cite{lepard2023} reconstructs 3D animals from images using a part-based neural representation. While applicable to animals with a different number of body parts, these methods do not reconstruct realistic fine-grained details, as the synthesis losses are based on matching silhouettes or image features. Moreover, they only reconstruct single animal instances. 
Methods exist to learn category-specific shape priors from images:
MagicPony \cite{Wu_2023_CVPR} learns models for horses, 3D-Fauna \cite{li2024learning} extends the approach to arbitrary quadrupeds. RAC \cite{Yang_2023_CVPR} learns category-level 3D models from video. GART \cite{lei2023gart} learns a subject specific model from monocular video.

\paragraph{3D Arboreal Trees Generation}
The modeling of trees and vegetation has a long history. Early approaches focused on modeling the branching structure, using fractals \cite{Aono1984_botanical, oppen1986_fractal}, grammars and particle systems \cite{jain2021_grammar} and L-systems \cite{Prusinkiewicz1990}, with the latter proved effective to modeling a large variety of realistic trees, given a set of production rules. Weber and Penn \cite{weber1995} define a procedural model that, instead of accurately modeling how trees grow, has a focus on the tree global geometry.
Using such systems is complicated, requires a lot of knowledge to define a non-intuitive set of parameters.
Recent methods exploit learning systems to easily define parameters, and automate the synthetic tree generation process. 
The recent DeepTree \cite{zhou2023deeptree} learns rules from traditional procedural methods, and define a network that is able to automatically grow trees taking into account environment constraints. %, without the need of a complex parameters definition. 
Lee et al.\cite{lee_latentL} train a neural network to generate parameters for procedural tree generation. None of these methods allow obtaining parameters from text, like we do. 
Li et al. \cite{li_learning_trees} grow tree branches using a multi-cylindrical shape, estimated from an image mask, as surface limit.

\section{Method}
\label{sec:method}

\subsection{Animal Model}
The \smalName parametric animal model is an extension of SMAL \cite{Zuffi:CVPR:2017}.
SMAL is defined by a triangular mesh template $\textbf{v}_{t}$, with $n_V$ vertices, a matrix $B$ of shape $3n_V{\times}n_B$ containing the $n_B$ basis vectors of a linear shape deformation space, a joint regressor $J_r$ that maps model vertices to a set of $n_J$ joint locations, and a skinning weight matrix $W$. 
An animal is generated, given shape parameters $\beta$ and pose parameters $\theta$, by first deforming the template into an intrinsic shape $\textbf{v}_{s}$, then applying Linear Blend Skinning (LBS) to rotate the body parts according to the given pose:
\begin{eqnarray}
\textbf{v}_{s} & = & \textbf{v}_{t} + B \beta^T \nonumber \\
\textbf{v} & = & LBS(\textbf{v}_{s}, \theta; W, J_r). 
\label{eq:model}
\end{eqnarray}

The linear shape space is learned with Principal Component Analysis (PCA) on a set of $41$ quadruped toy scans. 
The \smalName we introduce here is obtained by leveraging the training samples of SMAL, D-SMAL \cite{bite2023rueegg} and hSMAL \cite{li2021hsmal}.
We register the training horses of the hSMAL model plus additional horse toy scans to the SMAL topology, obtaining a set of $60$ registrations. We also add new species: Giraffe, Bear, Mouse, and Rat. We then learn an animal model on a total of $145$ animals. Note that D-SMAL defines dog breeds for the training samples, while in hSMAL the breed of the training horses is undefined.
After learning, we collect the set of shape variables for all the training samples, with their associated species or, in the case of dogs, breed name. This constitutes the training set for the \methodName animal shape prediction. 
\begin{figure}[t]
  \centering
\includegraphics[width=\linewidth]{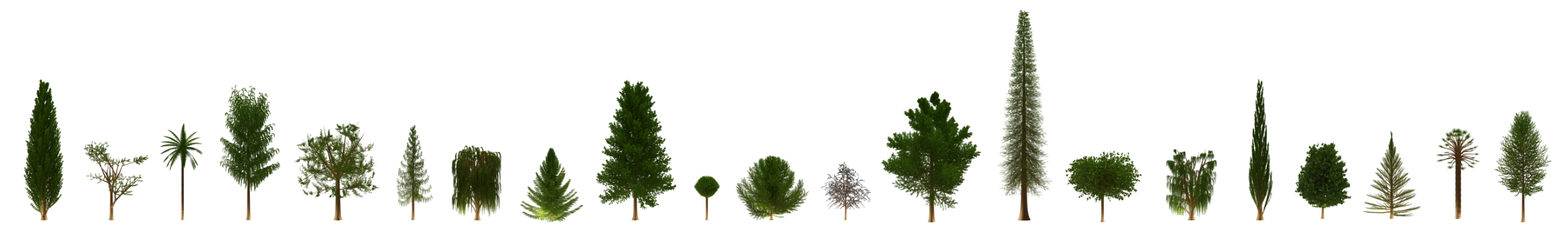}
  \caption{Training set for the tree network. From left: Poplar, Maple, Palm, Silver Birch, English Oak, European Larch, Weeping Willow, Balsam Fir, Black Tupelo, Sphere Tree, Black Oak, Hill Cherry, Sassafras, Douglas Fir, Apple, Willow, Cypress, Magnolia, Pine, Fan Palm, Quaking Aspen.
  }
  \label{fig:tree_training_set}
  \vspace{-0.2cm}
\end{figure}
\subsection{Tree Model}
The tree model corresponds to the TreeGen add-on for Blender \cite{treegen}.
Tree-Gen procedurally generates realistic 3D models of trees on the basis of the method proposed by Weber and Penn\cite{weber1995} and exploiting the Blender's Bézier curve system. The add-on supports saving the generated tree as a triangular mesh. 
The model generation is controlled by a set of parameters.
Some of the parameters are categorical, referring to a set of defined tree or leaf shapes, and some are numerical, controlling the branches and leaves density. In addition, ranges of variation for the numerical parameters are also defined, such that the add-on can generate diverse results for the same set of reference parameters.
Tree-Gen provides reference parameters labeled with the species name for a set of representative tree shapes. We add to the reference trees the Italian Cypress and Magnolia. This extended set of parameters and tree names constitutes the training set for the \methodName tree shape prediction (Fig. \ref{fig:tree_training_set}). 
\subsection{Text-to-Shape Model}
\begin{figure}[t]
  \centering
\includegraphics[width=\textwidth]{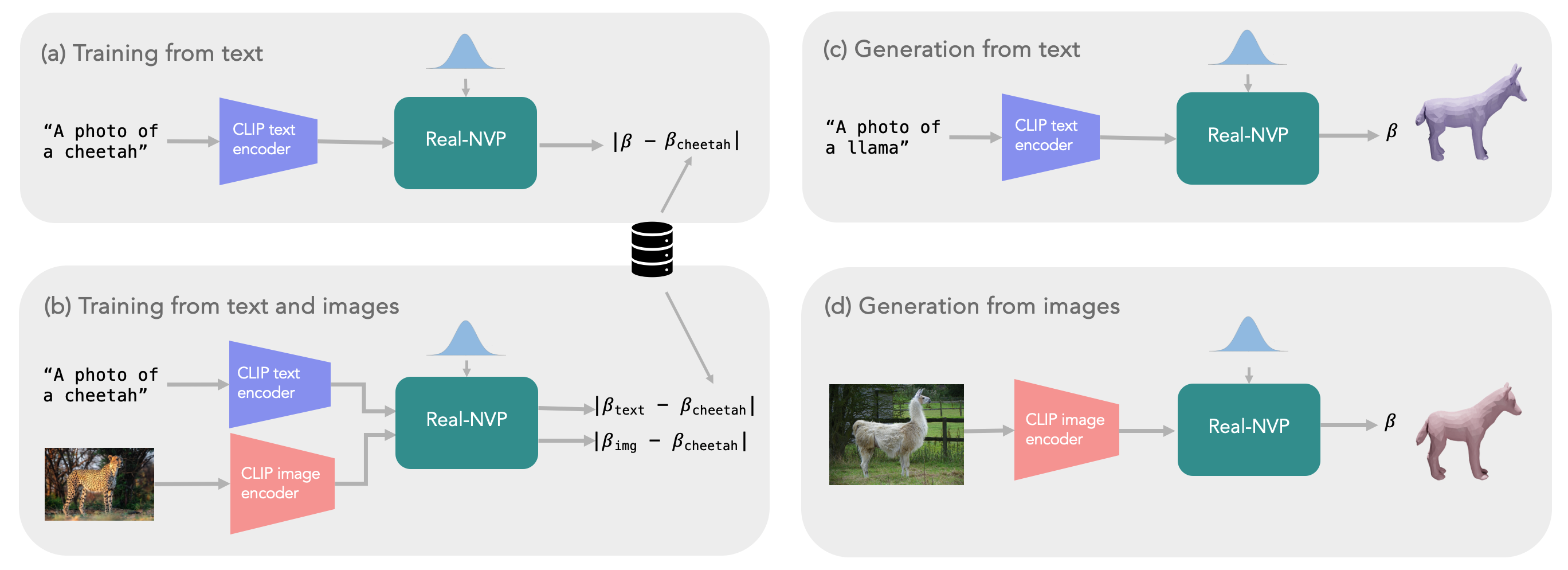}
  \caption{Network architecture. At training, we can consider only text as input (a), or also provide reference images (b), with about $3-10$ examples for each breed/species. At inference, we can query the text-only network with text (c), or the text-and-image network with images (d).
  }
  \label{fig:network}
  \vspace{-0.5cm}
\end{figure}

We base our approach on the real-valued non-volume preserving (\rnvpName) model \cite{Dinh2017realNVP}. \rnvpName is a generative probabilistic model specifically designed for high-dimensional and highly structured data.
Being formulated with a set of stably invertible transformations, and allowing exact and efficient reconstruction, \rnvpName is particularly suited for our task of latent space mapping with limited training data.
We summarize \rnvpName here.
Let $x\!\in\!X$ be an observed, high-dimensional variable, and $z\!\in\!Z$ a latent variable, with an associated simple prior distribution $p_Z$. Let $f$ be a bijection $f:X\!\rightarrow\!Z$, with $f^{-1}\!=\!g:Z\!\rightarrow\!X$. Using the change of variable formula, a model on $x$ can be defined as:
\begin{equation}
    p_X(x) = p_Z(f(x)) \Big| det \left( \frac{\partial f(x)}{\partial x^T} \right) \Big|,
\label{eq:rnvp}
\end{equation}
where the determinant is computed over the Jacobian of $f$. In order to generate samples from $p_X(x)$ one would first sample a latent variable $z$ from $p_Z$, then compute $x=g(z)$. Obtaining the density at $x$ requires computing the Jacobian (Eq. \ref{eq:rnvp}). 
Dinh et al. \cite{Dinh2017realNVP} introduce a convenient construction of $f$ using a set of bijective functions that are easy to invert. They formulate $f$ in a way that its Jacobian is a triangular matrix, allowing for the determinant computation as the product of the diagonal terms. Specifically, $f$ is obtained by stacking a set of \emph{Affine Coupling Layers}. Each coupling layer computes a transformation from the input $x\!\in\! \mathbb{R}^D$ to the output $y\!\in\! \mathbb{R}^D$ as follows:
\begin{eqnarray}
y_{1:d} = x_{1:d} \nonumber \\
y_{d+1:D} = x_{d+1:D} \odot \mathtt{exp} (s(x_{1:d})) + t(x_{1:d}),
\label{eq:rnvp2}
\end{eqnarray}
where $d\!<\!D$ and $s()$ and $t()$ are scale and translation functions that convert the input into a vector of dimension $D\!-\!d$. These transformations are easy to invert, and obtaining the Jacobian does not require computing derivatives for the scale and translation function \cite{Dinh2017realNVP}.
The partitioning of the input vectors can be modeled with a binary mask. In \cite{Dinh2017realNVP} two strategies are considered: checkerboard masking and dimension-wise masking. 
In \methodName we employ \rnvpName to model the conditional distribution of the shape parameters (either shape variables $\beta$ in \smalName or the parameters of the tree Blender add-on), given the CLIP encoding of the textual or visual input. Following \cite{sanghi2021clipForge}, we define the input variable $x$ in Eq. \ref{eq:rnvp2} as the concatenation between the CLIP encoding and the shape parameters. The output variable $z$ is a unit Gaussian distribution.
We adopt the \rnvpName model with important differences.
First, instead of considering a fix masking like in previous work \cite{Dinh2017realNVP, sanghi2021clipForge}, we consider trainable masks. 
Second, differently from the original formulation \cite{Dinh2017realNVP} and ClipForge \cite{sanghi2021clipForge}, %given our small data learning regimes, 
we seek for data reconstruction during training, employing a reconstruction loss, rather than a density estimation loss. This approach has been proved effective for training generative diffusion models \cite{tevet2023human}. We compare the different training losses in our ablation studies. As reconstruction loss, we use the $L1$ norm between predicted and ground truth shape parameters (See Fig. \ref{fig:network}). Note that, during development, we also considered a $L2$ loss, with poor results. 
Finally, we follow previous work in defining simple small networks to implement the scale and translation functions, namely two Multi Layer Perceptron (MLP) networks. Differently from previous work, we add two additional fully-connected layers that compress the hidden space of those functions. We found that this compression layer is necessary when learning the binary masks, while it hurts performance when the traditional masking approaches are considered.
We show the advantages of our design choices in our experiments.

\section{Experiments}

\label{sec:experiments}
\begin{figure}[t]
  \centering
\includegraphics[width=\textwidth]{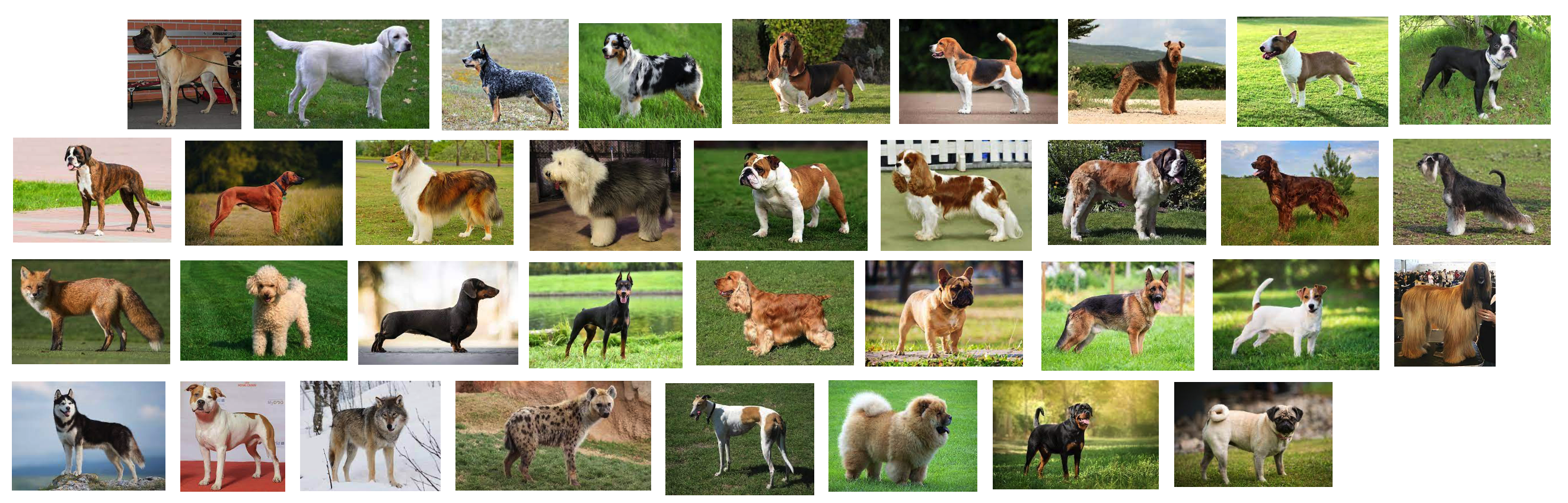}
  %\vspace{-0.6cm}
  \caption{Dog breeds. We verify that CLIP can discriminate the dog breeds in the D-SMAL training set by running a zero-shot classification test on the images above, which achieved $100\%$ accuracy.
  }
  \label{fig:dbreeds}
  \vspace{-0.5cm}
\end{figure}

\begin{figure}[b]
  \centering
\includegraphics[width=\textwidth]{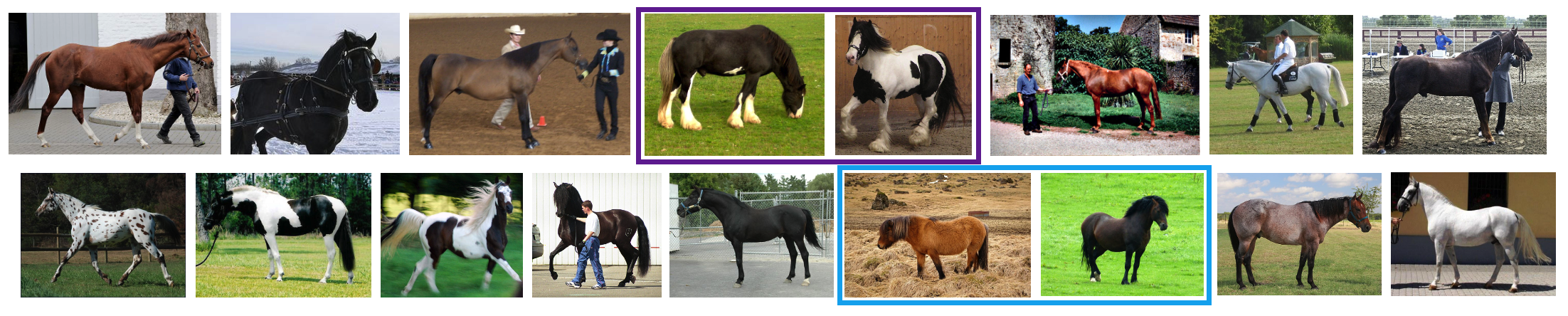}
   \vspace{-0.6cm}
   \caption{Horse breeds. We found with a zero-shot classification test that among the horse breeds above, CLIP can correctly recognize only for the Tinker/Shire horses (violet box) and the Icelandic/Welsh ponies (blue box).
  }
  \label{fig:hbreeds}
  \vspace{-0.5cm}
\end{figure}

We first verify that CLIP can understand and discriminate between the different dog breeds and tree species. We consider an image for each of the dog breeds in the D-SMAL model (see Fig. \ref{fig:dbreeds}), and perform zero-shot classification using the prompt "A photo of a <breed> dog". We found that CLIP can recognize all our training breeds.
Interestingly, the Chevalier King Charles Spaniel is correctly detected only if indicated as King Charles Spaniel.
We perform a similar experiment for our training tree species (Fig. \ref{fig:tree_training_set}) and a set of representative horse breeds (see Fig. \ref{fig:hbreeds}). 
We found that the most distinctive trees are correctly recognized, while the majority of the horse breeds cannot be identified, except for ponies and big horses. Therefore, we identify such cases in our animal training set, and assign corresponding labels, while the remaining horses are generically labeled as "Horse". 
\subsection{Implementation}
We implement the \methodName network in Pytorch.
We define a single network for both the animal and tree data, with similar training parameters, and the main difference being the dimension of the shape space.
The latent shape space for the animal network is the $145$-dimensional space of the \smalName model. The values of the shape variables are Gaussian distributed with zero mean and identity variance by construction.  
The latent space for the tree network corresponds to the parameters of the Blender add-on for tree generation.
We set the parameters that define the degree of randomness to zero, and we consider, to define the latent space, only parameters that vary across the reference species. This leaves a latent space with $60$ parameters out of the $105$ defined by Tree-Gen.
We center and normalize the variables by subtracting the mean and dividing by the standard deviation, such that the animals and tree parameters are defined with similar ranges.
We do not apply the centering and normalization to the categorical variables, that we represent instead with a one-hot encoding.
We consider $5$ affine coupling layers, the hidden space for the scale and translation networks has dimension $1024$, that we compress to $512$ with an additional layer.
We encode the text of the sentence "A photo of a <animal>" and "A photo of a <species> tree" for the animal and tree networks, respectively.
We train the animal and tree networks on the text and shape data for $6000$ epochs, which corresponds to a stabilization of the loss.  
We then train the same networks on text and images (Fig. \ref{fig:network} (b)). To do so, we download from the Web\footnote{https://commons.wikimedia.org/} a set of images, between $3$ to $10$ for each tree/animal species or breed, and create training tuples composed by the CLIP image encoding and parameters. The training data is larger than previously, and we train the networks for $3000$ epochs, which corresponds to a stabilization of the loss.
Batch size is $16$. We use the Adam optimizer with a learning rate that varies from $1e\!-\!4$ to $1e\!-\!6$. We use CLIP ViT-B/32-LAION-2B \cite{ilharco2021_openclip}.
\begin{figure}
  \centering
  \includegraphics[width=\textwidth]{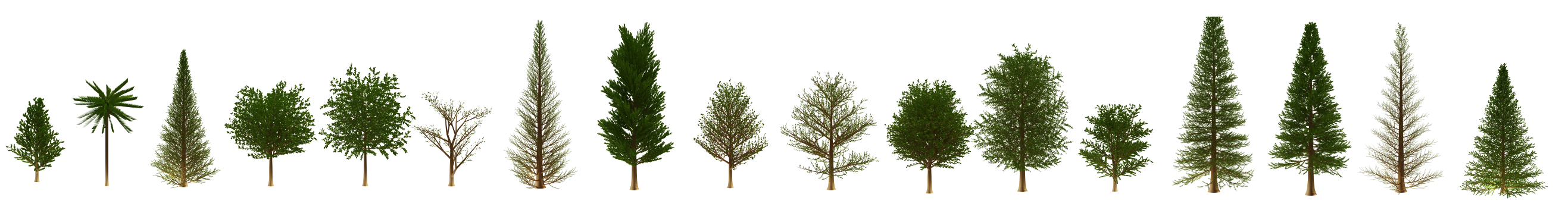}
   \vspace{-0.6cm}
  \caption{Tree prediction from text. The generated tree species are, from left: Gingko, Coconut, Cedar of Lebanon, Fig, Cocoa, Bigleaf Maple, Deodar Cedar, Eucalyptus, Tulip, Oak, Banyan, American Elm, Acer, Coast Redwood, Sequoia, Western Red Cedar, White Spruce. None of these species is in the tree training set.
  }
  \label{fig:trees_results}
\end{figure}
\begin{figure}[p]
  \centering  \includegraphics[width=\textwidth]{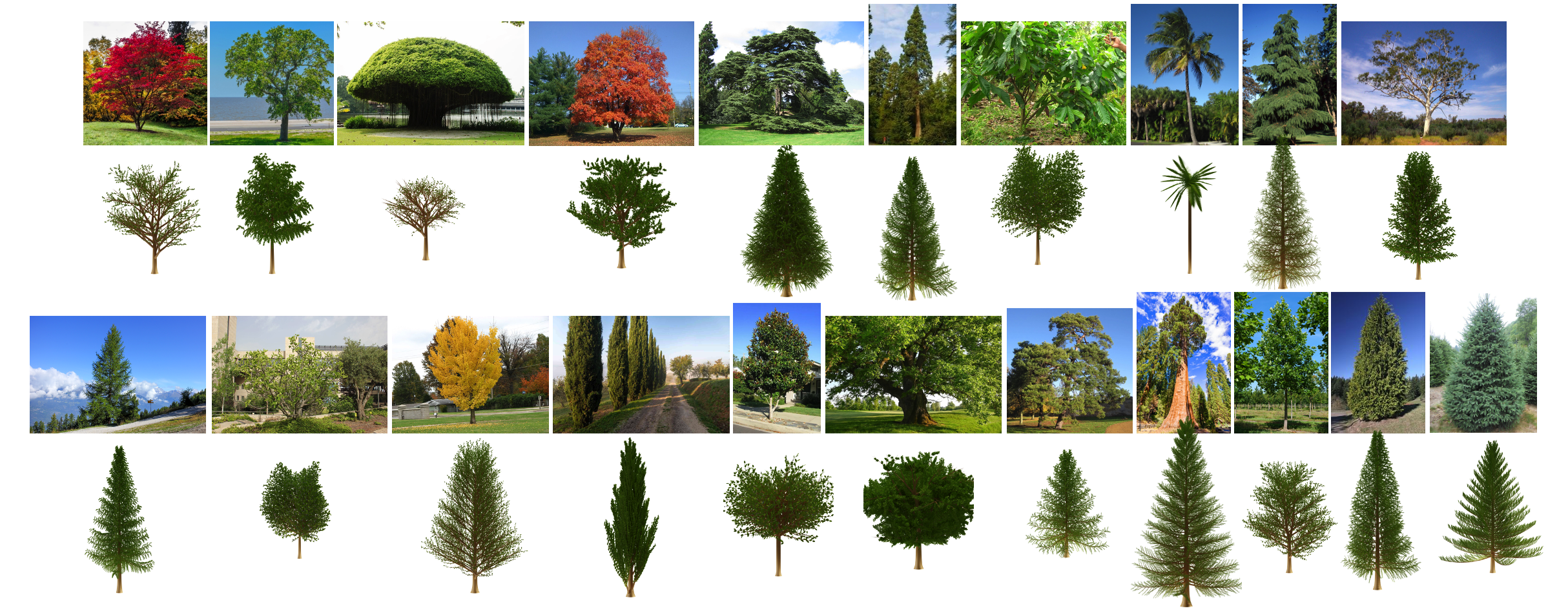}
   \vspace{-0.6cm}
  \caption{Tree prediction from images. For each row: the input image and below the generated tree. Note that we do not predict the tree colors, and we show all the trees with an average green.
  }
  \label{fig:trees_results_images}
\end{figure}
\vspace{-0.5cm}
\subsection{Evaluation}
We evaluate our \methodName method in two settings: interpolation and generalization.

\begin{figure}
  \centering
  \includegraphics[width=\textwidth]{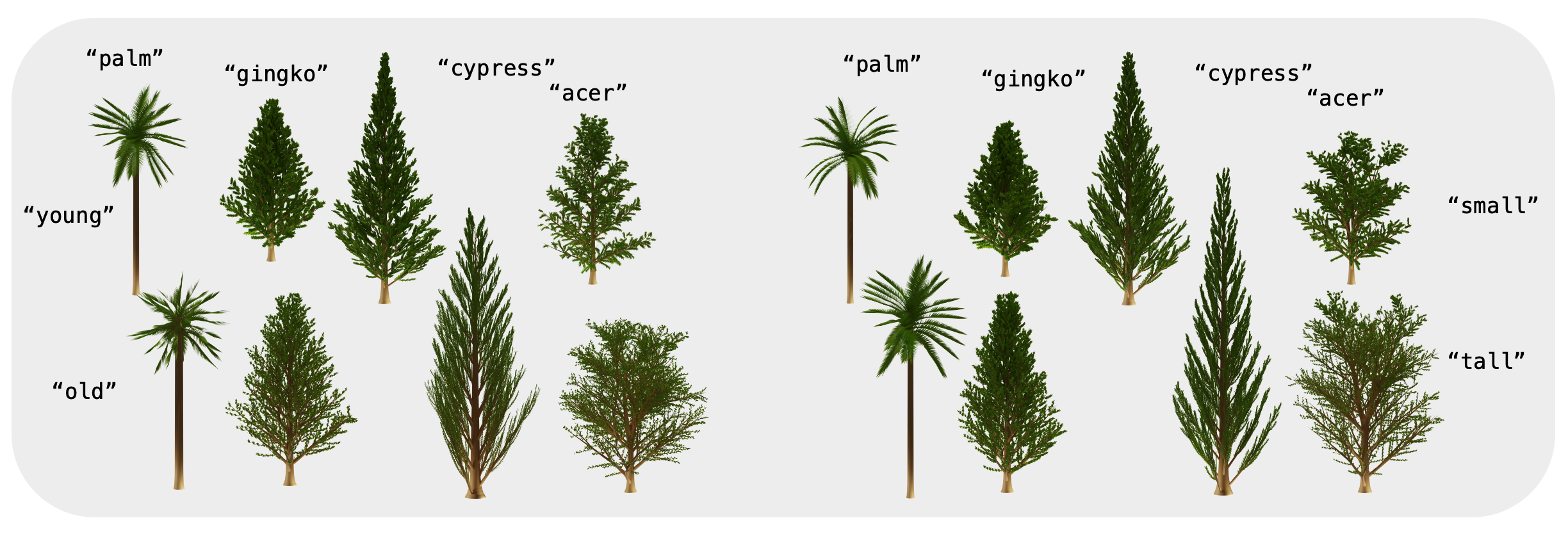}
   \vspace{-0.6cm}
  \caption{Age and size interpolation for trees. Palm and Cypress are in the \methodName training set, while Gingko and Acer are unseen species. Here the query is "A photo of a <age> <species> tree".
  }
  \label{fig:trees_interp}
   \vspace{-0.2cm}
\end{figure}

\begin{figure}
  \centering
  \includegraphics[width=\textwidth]{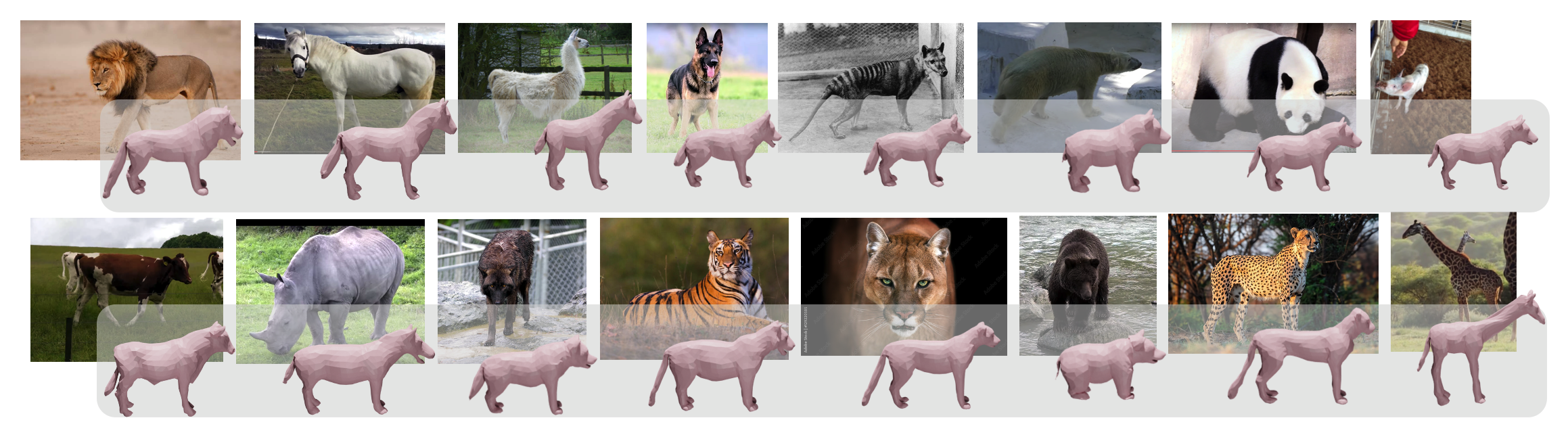}
   \vspace{-0.6cm}
  \caption{Animals prediction from images. The Horse, Dog, Thylacine, Polar Bear, Panda, Pig, Cow, Rhino and Bear images are taken from \cite{Zuffi:CVPR:2018}. We replace their green screen images with natural images for the Lion, Tiger, Cougar and Cheetah. 
  }
  \label{fig:animals_from_images_results}
  \vspace{-0.5cm}
\end{figure}

\begin{figure}
  \centering
  \includegraphics[width=\linewidth]
   {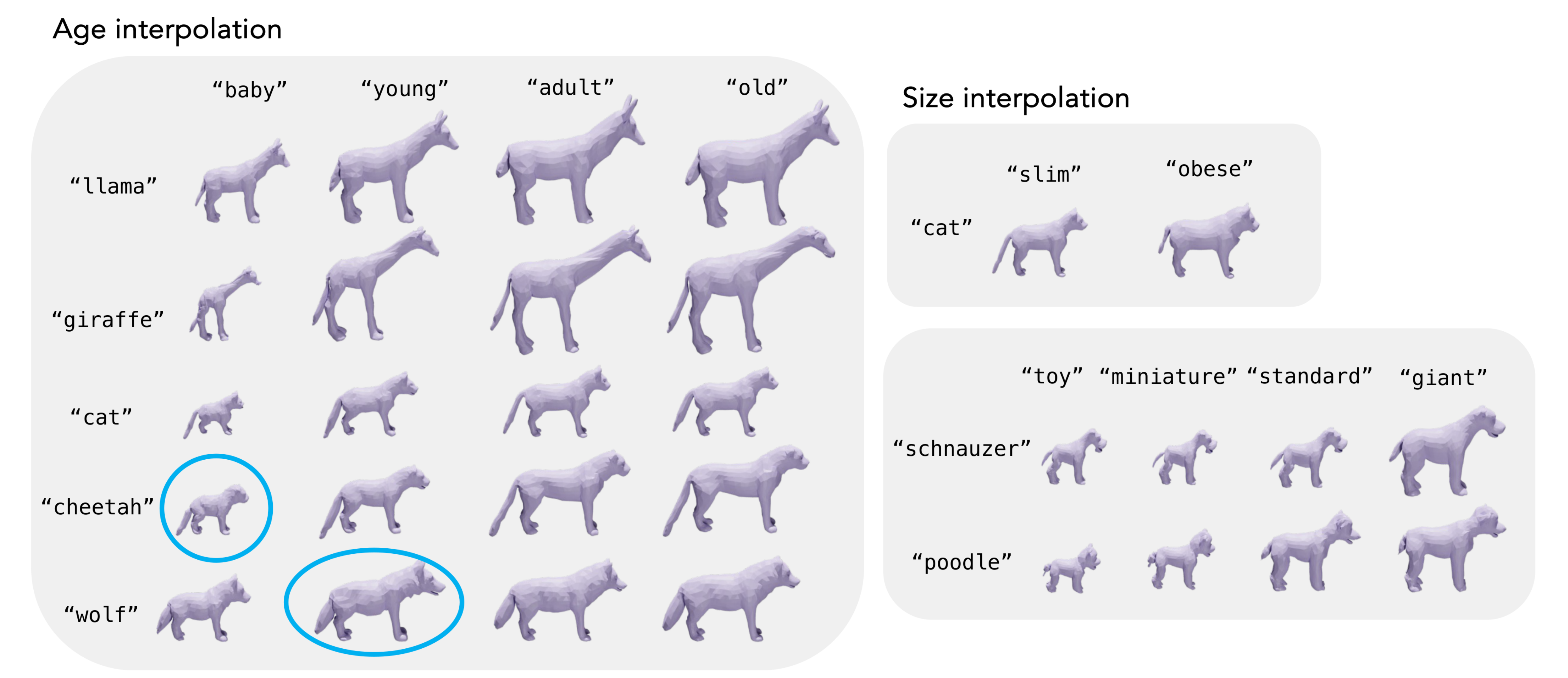}
   \vspace{-0.6cm}
   \caption{Age interpolation (left) and size interpolation (right). The circles indicate the animals that are present in the training set as "Baby Cheetah" and "Young Wolf". Giraffe, Cat, and Wolf are in the training set without attributes, while the Llama is not. 
  The small and large Poodles are present in the training as 3D shapes, but their text attribute is "Poodle". Only one shape example for the Schnauzer is present, named "Schnauzer". Note how we can recover the different Poodle breed size variations. For the Schnauzer, the actual breed variations are Miniature, Standard and Giant.}
  
  \label{fig:interpolation}
  \vspace{-0.5cm}
\end{figure}

\begin{figure}
  \centering
  \includegraphics[width=\textwidth]{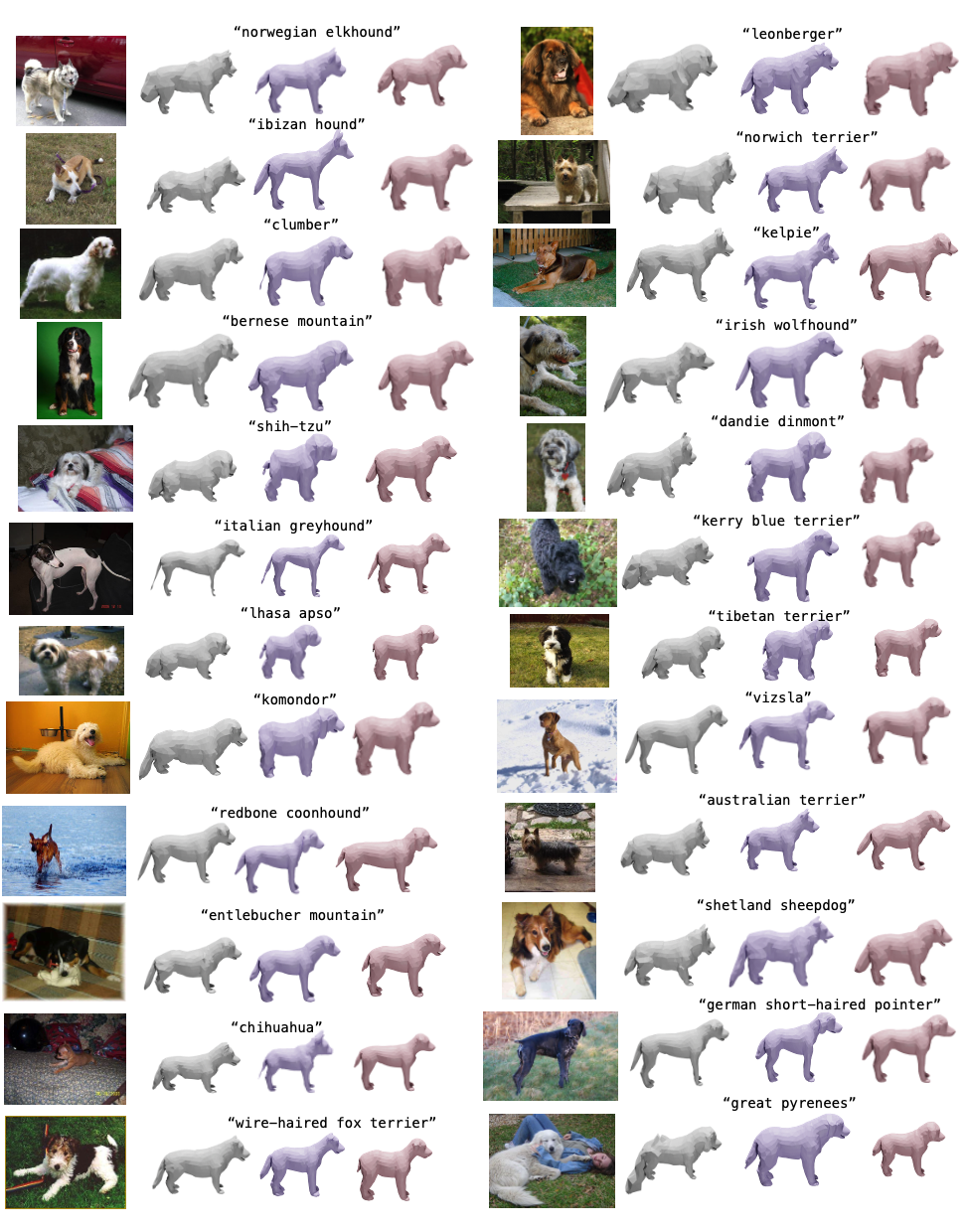}
   \vspace{-0.6cm}
   \caption{Comparison with BITE\cite{bite2023rueegg}. Randomly chosen images from the StanfordExtra testset. From left: input image, BITE result rendered in a natural pose (gray), \methodName result with textual input (purple), \methodName result with image input (red). For BITE and \methodName with text input, we use the breed label to rotate the ears. For \methodName from images, ears are down by default. None of these breeds are in the \methodName training set.
  }
  \label{fig:bite_comparison}
\end{figure}

\noindent\textbf{Interpolation}. We consider as interpolation task the prediction of new breeds for the dog class. In nature, dogs of different breeds can mix, and many breeds have been created by mixing existing ones \cite{Parker2017}. We argue that, given the large number of breeds included in the model, it is likely that new breeds shapes can be generated by interpolation in the space of dog shapes, even if it is true that there could be unseen breeds with specific shape features not seen during training. 
We qualitatively demonstrate interpolation by generating dog breeds in comparison with BITE\cite{bite2023rueegg} (Fig. \ref{fig:bite_comparison}).
We also show interpolation for age and size. We query for "Giant Schnauzer", "Standard Schnauzer", "Miniature Schnauzer" and "Toy Schnauzer", and similarly for the Poodle. Note in Figure \ref{fig:interpolation} how the network correctly predicts the scale of the different varieties of the breeds (it is worth noting that for the Schnauzer, the breed varieties are only Giant, Standard and Miniature). We then investigate if \methodName can interpolate shapes and age-dependent features by querying for "Baby", "Young", "Adult" and "Old" animals. Figure \ref{fig:interpolation} shows the results for seen and unseen species.  
Figure \ref{fig:trees_interp} shows an analogous analysis for trees.
We quantitatively compare the dog breed predictions from textual input with BITE \cite{bite2023rueegg} with a perceptual study. For each breed in the StanfordExtra testset \cite{biggs2020wldo}, we generate a 3D dog, and compare with the dog reconstructed by BITE on a randomly selected image of the same breed. We let Amazon Mechanical Turk workers to judge which method better represents the dog breed in the picture. On the whole set, BITE outperforms \methodName with $971$ vs $884$ votes, as confirmed by a binomial test, with a p-value of $0.02$ (BITE better than \methodName). We noticed that the task favors BITE when the subject in the image is a puppy, given \methodName used without age input generates an adult subject. By removing from the evaluation the images with baby dogs, we obtain votes of $830$ (BITE) vs. $850$ (\methodName), with a p-value of $0.3$ (\methodName better than BITE), indicating the ability of \methodName to faithfully generate a large variety of breeds.
\begin{figure}
  \centering
  \includegraphics[width=\textwidth]{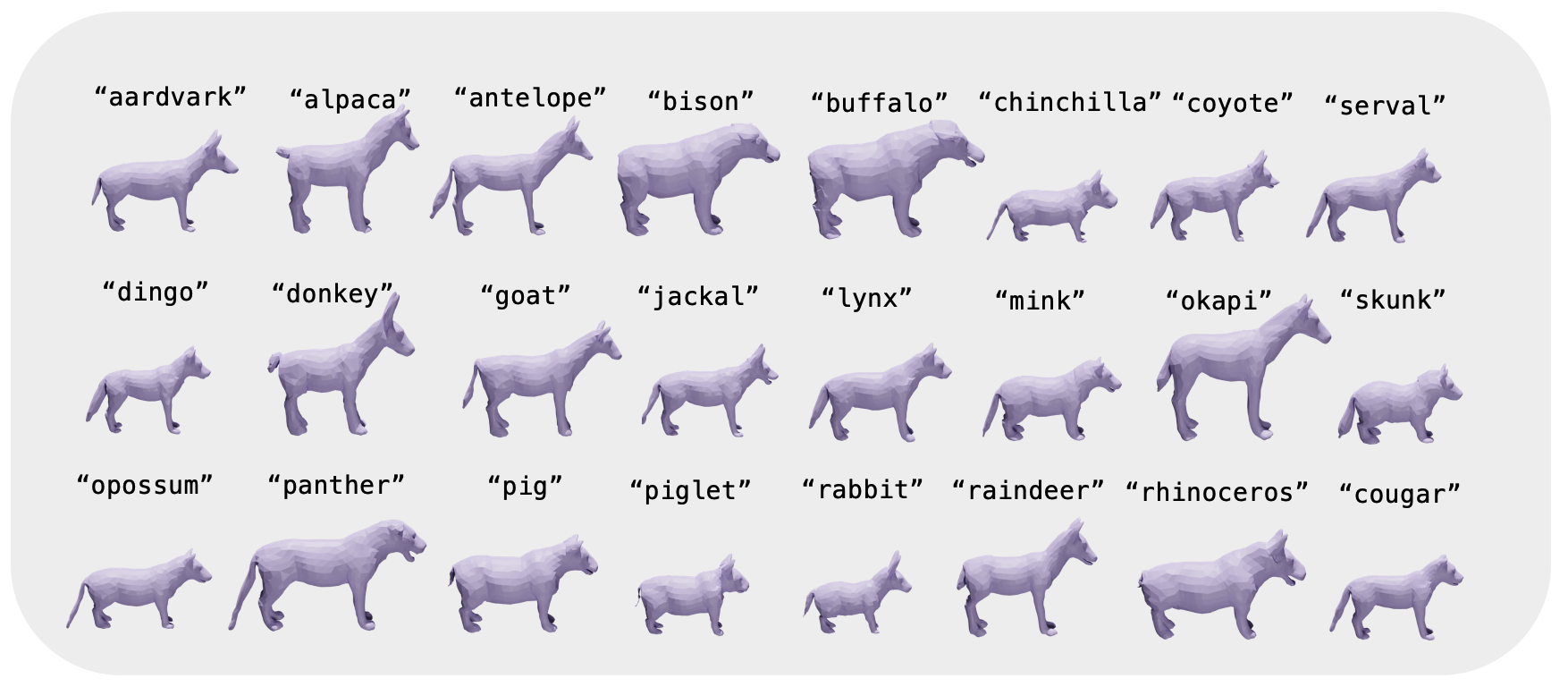}
   \vspace{-0.6cm}
  \caption{Animals prediction from text. We generate species that are not present in the \smalName and \methodName training sets. The image shows the actual model size. 
  }
  \label{fig:generalization}
\end{figure}
\begin{figure}
  \centering
  \includegraphics[width=\textwidth]{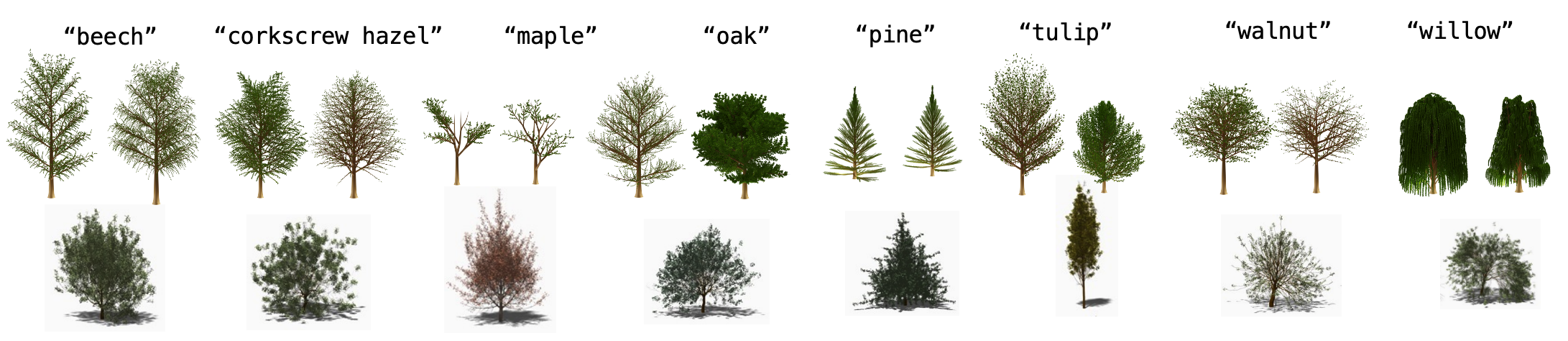}
   \vspace{-0.6cm}
   \caption{Comparison with DeepTree. We show predicted trees from text (top), compared with DeepTree (bottom, images taken from  \cite{zhou2023deeptree}). For each predicted pair: (left) network trained only on text, (right) network trained on text and images. 
  }
  \label{fig:deeptree}
\end{figure}
\noindent\textbf{Generalization}. In order to test generalization, we prompt the model for creating new quadruped species. Figure \ref{fig:generalization} shows examples of generation from text.
 We show examples of reconstructed novel trees from textual and image input in Figure \ref{fig:trees_results} and Figure \ref{fig:trees_results_images}, respectively.
 Finally, Figure \ref{fig:animals_from_images_results} shows examples of generation of animals from images, where many are taken from \cite{Zuffi:CVPR:2018}, for comparison. Here the unseen animals are the Llama, Thylacine, Panda, Pig, Rhino, Cougar. Figure \ref{fig:deeptree} shows a comparison with DeepTree \cite{zhou2023deeptree}.
\begin{table}[t]
    \centering
    \begin{tabular}{l cccc }
    \toprule
        \multicolumn{1}{c}{} &\multicolumn{3}{c}{CLIP-based Comparison: \% of votes}\\
        \multicolumn{1}{c}{} & All & p-value & Dogs & Other Species  \\%(all)\\ 
        \hline
        A. Check  vs. Dims  & 61:39 & 0.19 & 68:32 & 43:57 \\ 
        B. Dims vs. Dims + Comp. & 52:48 & 0.47 & 51:49 & 54:46  \\ 
        C. Check vs. Check. + Comp. & 63:37  & 0.20 & 64:36 & 60:40  \\ 
        D. Learn + Comp. vs. Learn & 61:39 & 0.19 & 59:41 & 69:31  \\ 
        E. Learn + Comp., 145 vs. 40 & 50:50 & 0.58 & 48:52 & 54:46 \\  
        F. Learn + Comp. vs. Dims & 62:38 & 0.13 & 68:32 & 46:54  \\ 
        G. Learn + Comp vs. Check & 54:46 & 0.38 & 53:47 & 57:43 
        \\ 
        H. Learn + Comp, density loss & 86:14 & 1.24e-7 & 86:14 & 86:14 
        \\ 
        \hline        
    \end{tabular}
\caption{Ablation results. Comparison between different networks. Check is checkerboard masking, Dims is dimension-wise masking, Comp is hidden space compression, Learn is learned masks. (E) compares the Learn + Comp network with 145 (default) versus 40 shape parameters.
The table show that the best performance on the whole testset is for the network with learned mask and compression (D, F, G). When training with also a density loss \cite{Dinh2017realNVP}, performance degrades significantly.} 
\label{tab:ablation}
\end{table}

\subsection{Ablation Studies}
We perform our ablation studies on the animal model. We use CLIP for evaluation, as we found that CLIP can successfully classify animals and dog breeds in particular, allowing quantitative testing on a larger set of cases. 
We perform ablation studies to evaluate: the effect of learning the binary masks in \rnvpName; the effect of training with a density loss; the effect of adding the compression layer in the scale and translation functions. We also compare with reducing the shape space dimension from $145$, the space of the \smalName model, to a dimension of $40$, approximately matching the dimension of the dogs and horses single SMAL models \cite{bite2023rueegg}\cite{li2021hsmal}.
We generate a set of $122$ animals, none of them present in the \smalName model training set. This selection covers most of the common quadrupeds, and several unseen dog breeds. 
We query the network with the sentence "A photo of a <animal name>", where <animal name> is either a quadruped species or dog breed. We then render the predicted 3D models in grayscale, in order to prevent any color bias. 
 Given the networks can predict different animal sizes, we consider bounding boxes. %
We render the animals to maximize visibility of their profile. We found that the lateral view is the most informative, while adding further views gave inconsistent results.
We perform paired comparison between different networks by testing, for each animal, which of the two networks predictions, encoded in CLIP, is closest to the CLIP encoding of the animal name. This corresponds to a CLIP "vote".
Note that, even if we base our method on CLIP, we believe it is appropriate to use CLIP for the ablation studies, as we are comparing different architectures, under the same conditions.
Results are reported in Table \ref{tab:ablation}. Our ablation studies confirm that the network with learned masks and compression of the hidden space for the scale and translation networks provides the best performance on the whole testset. 

\section{Conclusion}
\vspace{-2mm}
We have addressed the problem of generating 3D objects from text and images using parametric 3D models. Inspired by recent work on learning multimodal latent spaces, we use language to control the selection of the 3D models parameters. We make the hypothesis that using language we can achieve interpolation and generalization in parametric shape spaces.
We demonstrate our hypothesis on two different 3D generative models: on a novel differentiable 3D parametric shape model for animals, that extends previous models with new training samples and species, and on a non-differentiable model for trees, represented by a Blender add-on. Our qualitative and quantitative experiments confirm our hypothesis. The proposed \methodName is the first system that allows generating rigged 3D animals and trees with a simple text prompt.
\noindent\textbf{Acknowledgements}. We thank Tsvetelina Alexiadis, Taylor McConnell and Tomasz Niewiadomski for the huge help in running the Amazon Mechanical Turk evaluation. We also thank Charlie Hewitt for making his tree generation method available and the authors of \cite{sanghi2021clipForge} for sharing their code.

% ---- Bibliography ----
%
\bibliographystyle{splncs04}
\bibliography{egbib}
\end{document}